\documentclass[onecolumn,11pt]{IEEEtran}

\def\No{{$-$}}
\def\Yes{\checkmark}

\usepackage{times}
\usepackage{graphicx}
\usepackage{amsmath,amsthm}
\usepackage{amssymb,empheq}
\usepackage{latexsym}
\usepackage{epstopdf}
\usepackage{url}
\usepackage{multirow}

\usepackage{sverb}

\usepackage[ruled,linesnumbered]{algorithm2e}
\usepackage{algorithmic}

\usepackage[usenames]{color}
\definecolor{shadecolour}{gray}{0.4}

\graphicspath{{./}{./Fig/}{./fig/}{./Figs/}{./figs/}}

\newcommand{\BW}{{\mathbf{W}}}

\newcommand{\Bx}{{\mathbf{x}}}

\renewcommand{\Lambda}{\varLambda}

\newcommand{\eg}{e.g.}

\newcommand{\etc}{{etc}}
\newcommand{\etal}{{et al.}}

\begin{document}


\title{Face Image Classification by Pooling Raw Features}

\author{
          Fumin Shen,
         Chunhua Shen 
         and
         Heng Tao Shen
\thanks
{
    F. Shen is with School of Computer Science and Engineering, University of Electronic Science and Technology of China, Chengdu 611731, P.R. China 
    (e-mail: fumin.shen@gmail.com; shenhengtao@hotmail.com).
   Part of this work was done
   when the first authour was visiting The University of Adelaide. 
  Correspondence should be addressed to F. Shen.
   }
   
   \thanks
{
    C. Shen is with The Australian Center for Visual Technologies, and School
    of Computer Science at The University of Adelaide, SA 5005, Australia (e-mail: chunhua.shen@adelaide.edu.au). 
    }
    \thanks
    {
    H. T. Shen is with School of Information Technology and Electrical Engineering, The University of Queensland, Australia and School of Computer Science and Engineering, University of Electronic Science and Technology of China (E-mail: shenht@itee.uq.edu.au).
    }

}

\maketitle
\begin{abstract}
We propose a very simple, efficient yet surprisingly effective feature extraction method for face
recognition (about 20 lines of Matlab code), which is mainly inspired by spatial pyramid
pooling in generic image classification. We show that features formed by simply pooling local
patches over a multi-level pyramid, coupled with a linear classifier, can significantly
outperform most recent face recognition methods. The simplicity of our feature
extraction procedure is demonstrated by the fact that no learning is involved (except PCA
whitening).
%
%
We show that, multi-level spatial pooling and dense extraction of multi-scale patches play
critical roles in face image classification. The extracted facial features can capture strong
structural information of individual faces with no label information being used.
We also find that, pre-processing  on local image patches such as contrast normalization
can have an important impact on the classification accuracy.
In particular,
on the challenging face recognition datasets of FERET  and LFW-a, our method improves previous best results by
more than 10\% and 20\%, respectively. 
\end{abstract}

\section{Introduction}
\label{SEC:Intro}

Face recognition has been a long-standing research topic in computer vision.
It remains a challenging problem despite the extensive research effort spent on this problem.
These challenges are typically
 caused by various intra-class variations (\eg, face expressions, poses, ages, image
contaminations, etc.), or lack of sufficient training samples \cite{li2011handbook}. To achieve encouraging
recognition accuracies, it is believed that the key is to find
effective and robust representations of
facial images.
Conventional face image features in the literature include eigenfaces \cite{Turk91},
Fisher-faces \cite{FisherLDA}, Laplacian faces \cite{Laplacianfaces05}, \etc.
These features work well on certain tasks. However, they often cannot cope with the aforementioned problems
well such as image occlusions, large pose variations, \etc.

Recently, sparse representation based face classification has shown encouraging success
\cite{Wright09,yang2011fisher}. Different from  previous methods, which
make decisions based on  standard classifiers (\eg, $ k $-nearest-neighbor or support vector machines),
     the classification rule of these methods is based on the minimum representation
error of the probe image over a set of training samples or a dictionary learned from
training images. As a typical example, the sparse representation based classifier (SRC)
\cite{Wright09} linearly represents a probe image by all the training images under
the $L_1$ norm based sparsity constraint. The success of SRC has induced a few sparse
representation based approaches \cite{yang2011fisher,ESRC2012,dengdefense2013,RSC11},
which achieve state-of-the-art performance on face recognition with image corruptions
\cite{Wright09}, face disguises \cite{yang2011fisher,RSC11} and small-size training
data \cite{dengdefense2013}.

Compared to these methods mainly working on holistic features,
researchers found that  face recognition
based on local features or patches is often more robust to mis-alignment and occlusions.
For example, histograms of local binary patterns
(LBP) \cite{FR_LBP06}, Gabor features \cite{GaborWavelet93,zhang2007histogram,zou2007comparative} and their fusions \cite{tan2007fusing} have  proven to improve
the robustness of face recognition systems. 
 Another simple way to extract local features
is the modular approach, which first divides the entire facial image into several blocks
and then aggregates the classification outputs that were made independently on each of these local
blocks \cite{Wright09}. One of the widely-used aggregation strategies might be voting
methods \cite{kumar2011maximizing,kumar2012trainable,Wright09}. Recently, Zhu \etal
\cite{zhu2012multi} proposed a linear representation based method, which fuses the
results of multi-scale patches using boosting.

Local features or image patches has also been widely used in generic image
classification. The bag-of-features (BOF) model, which represents an image as a
collection of orderless local features, is the standard approach in image classification
\cite{grauman2005pyramid,SPM2006,yang2009linear}.
%
%
The spatial pyramid matching (SPM) method in \cite{SPM2006,yang2009linear}  has made a remarkable success in image
classification by explicitly incorporating spatial geometry information into classification.
SPM divides an image to increasingly fine spatial grids at different scales
with $2^l \times 2^l$ cells at the $l^{th}$ level.
Codes by encoding the local features (\eg, SIFT features) within each spatial cell over a
pre-trained dictionary  are pooled by average pooling or max-pooling.  Despite
the success of the spatial pooling model in image classification, it has rarely been used
in face recognition problems.

\begin{figure}[t!]
\centering
\includegraphics[width=0.4\textwidth]{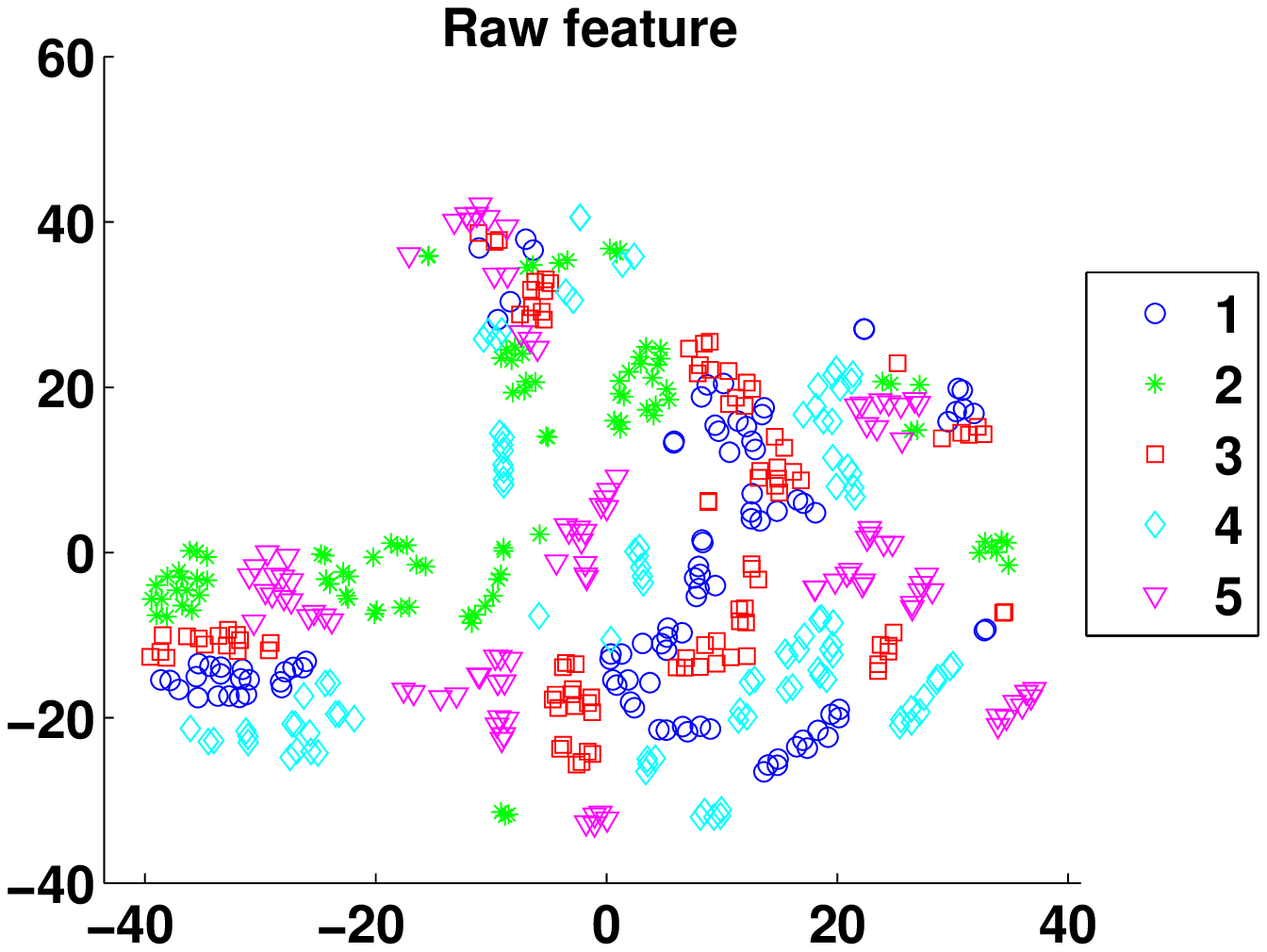}\;\;\;\;
\includegraphics[width=0.4\textwidth]{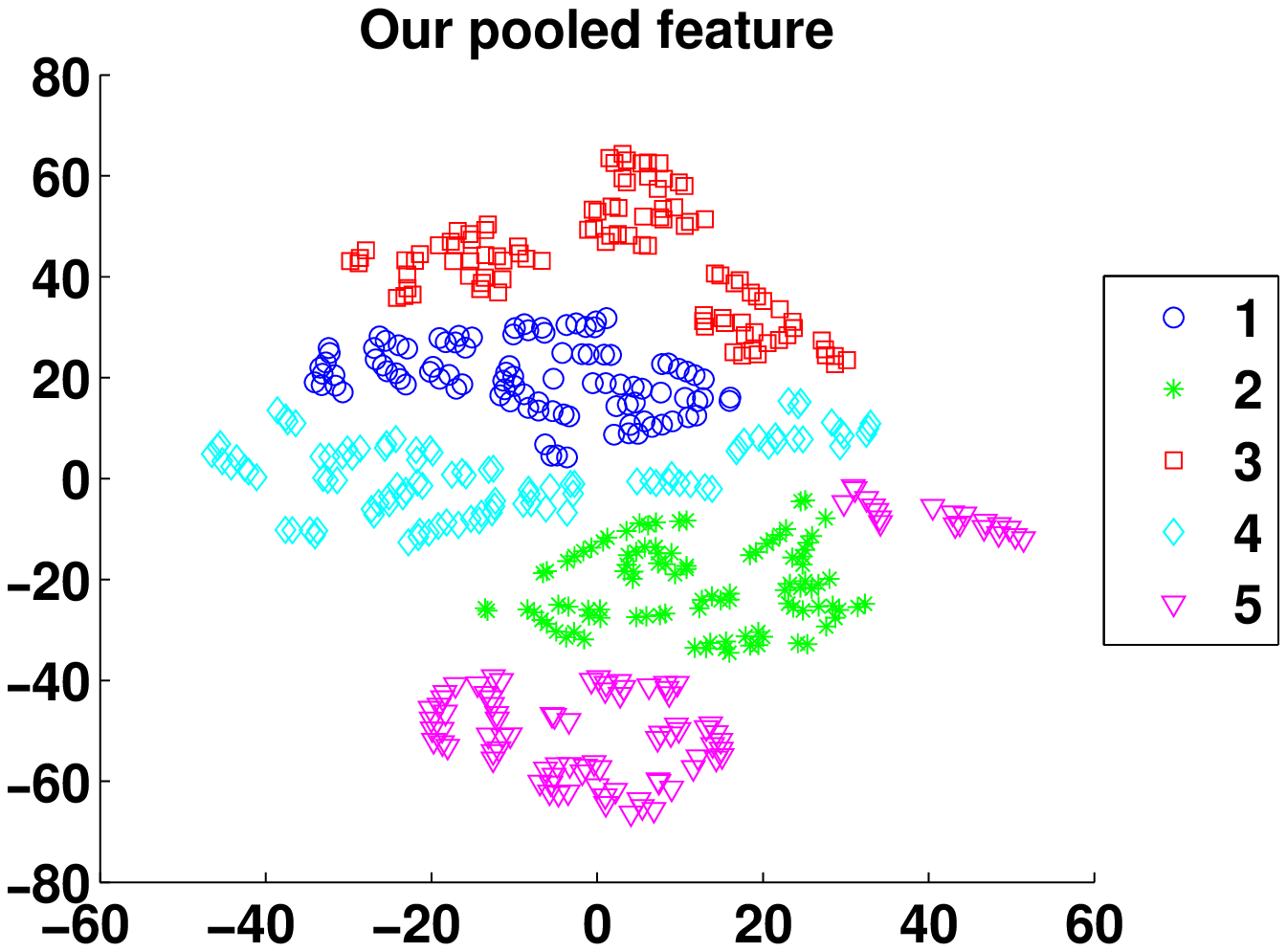}
\caption{Visualization of raw pixel features (top) and our pooled features (bottom) of face images
by the unsupervised embedding method t-SNE \cite{tSNE2008} into 2D.
    Here the first 5 classes of the CMU-PIE dataset \cite{PIE} are used, with each class
    having 100 face images. Data points belonging to the same class are marked with the same color and
    marker. Note that no label information is used here. We can clearly see that using our feature representations,
    faces in the same class tend to be grouped into a cluster and different classes are much better separated. Best viewed in color.}
\label{Fig:visual}
\end{figure}

Mainly inspired by the spatial pyramid pooling method, here we propose a simple feature
extraction method for face recognition. Instead of pooling the encoded features over a
pre-trained dictionary, {\em we compute the facial features by performing pooling directly
on the local patches, which are densely extracted from the face image}. In other words,
our pipeline removes the dictionary learning and feature encoding from the
standard unsupervised feature learning based generic image classification pipeline.
As a result, our feature extraction method is very simple, efficient and can be implemented in less than 20
lines of Matlab
code. A surprising result is that, however, coupled with a linear classifier,  our method outperforms
almost all the most recently reported methods by a large margin on several benchmark face
recognition datasets. The extracted facial features are shown to have strong structural
information of individuals without using label information. Fig.~\ref{Fig:visual}
illustrates an example.

In summary, the contributions of this work mainly include:
\begin{itemize}

\setlength{\itemsep}{-1mm}
\item[1.]

We propose a new facial feature extraction method based on spatial pooling of local patches and apply it to
face identification problems.
For the first time, we advocate directly applying pooling on local patches. Note that we directly work on raw pixels,
rather than local image features such as SIFT.

\item[2.]

The proposed facial feature extraction method is very simple yet extremely effective, which is
shown to outperform recently reported  face recognition methods on four tested public
benchmark databases. Accuracy improvements of over 10\% and 20\% are obtained by our
method over the previous best results on the FERET and LFW-a datasets, respectively.
The proposed method also works  well on the small training size, and in particular the single-sample-per-person (SSPP) face recognition problems.

\item[3.]
Compared to other local feature (\eg, LBP, Gabor feature) or local patch (\eg, BOF model) based methods, the proposed pooling method is much simpler, and achieves close or even better results.

\end{itemize}

Moreover, through a thorough evaluation, we show that  dense extraction of multi-scale patches
and spatial pooling over a deep pyramid play critical roles in achieving high-performance
face recognition. We also show that, the pre-processing operations such as contrast
normalization and  polarity splitting can have important impacts on the classification
performance as well.

\section{Our method}
\label{SEC:Method}

In this section we present the details of the proposed method.

\subsection{Local patch extraction and pre-processing}
\label{SEC:local}

Without loss of generality, suppose that a facial image is of $d \times d$ pixels.
As the first step,
we extract overlapped local patches of size $r \times r$  pixels with a step of $s$ pixels.
Set $l = \lfloor \frac{d - r}{s} + 1 \rfloor$,
then each image is divided into $l \times l$ patches.
Let each local patch be a
row vector $\Bx$. We then perform normalization on $\Bx$ as:
$
\hat{x}_i = {(x_i - m)}/{v}, 
$
where $x_i$ is the $i^{\rm th}$ element of $x$, and $m$ and $v$ are the mean and standard deviation of elements of $\Bx$.
This operation contributes to local brightness and contrast normalization as in
\cite{coates2011analysis}.

With each patch extracted from all training images contrast normalized,
an optional step is to conduct
dimensionality reduction on these patches by principal component analysis (PCA)
\cite{Turk91}. One of the benefits by doing this is de-noising. Of course by applying PCA, we also reduce
the dimensionality of the final feature. Later, in the experiments we  show that transforming the patches to a low dimensional (\eg, 10D, to capture 90\% of
the energy) eigenspace is typically sufficient.

It has been shown in object recognition that,
the simple absolute value rectification \cite{jarrett2009best} or polarity splitting
\cite{coates2011importance} on the extracted features before pooling often improves the
classification performance.
Here we apply the polarity splitting in our algorithm directly on the normalized patches:
we split the positive and negative components of each patch $\Bx$ into separate parts:
$\max\{\Bx, 0\}$ and $\max\{-\Bx, 0\}$.
These two parts are then concatenated to form a vector which is twice as long as the original vector.

\subsection{Spatial pyramid pooling}

Then we come to the most important part of our algorithm: spatially pooling the
pre-processed patches. Different from all the feature coding methods
\cite{coates2011importance,yang2009linear}, which conduct spatial pooling on the encoded features over
a pre-trained dictionary, {\em we propose to perform pooling directly on the local
patches.} Suppose that we set the pooling pyramid of $L$ levels to
$\{c_1, c_2,  \cdots, c_L \}$, such that the grid at level $l$ have $c_l$ cells at each dimension.
Then we have in total $\sum_{l = 1}^{L} c^2_l$ pooling cells for a 2D image.

The dimension of the processed patches can be very low, which makes it possible to apply a much
larger number of pooling levels.
As an example, after applying PCA, the dimension of each local patch
is about 10. While for
unsupervised feature learning or bag-of-features based methods \cite{coates2011importance,yang2009linear},
the encoded feature dimension for each extracted local feature is the same as
the dictionary size, which is
usually a few thousand---two orders of magnitude larger.
We show in the next section that, {\em for the proposed method, a large
number of pooling pyramid levels are critical to achieve a high accuracy in face recognition}.

Those popular pooling methods, \eg, average pooling and max-pooling,
can be used in our algorithm.
They can easily be performed by the average- or max-operation along
each normalized pixel of the patches (again, not on the encoded features but directly on the pixels),
in the pooling cell:
\begin{align}
f_{ i } &=  \textstyle \sum_j x_i^{ (j) }  / m,    \tag{average pooling}\\
f_{ i } &=  \max_j x_{i}^{(j)}               \tag{max pooling}.
\end{align}
Here $x_i^{(j)}$ is the $i^{\rm th}$ element of the $j^{\rm th}$ local patch in the current pooling cell.
$m$ is the number of patches in the pooling cell  and
$ {\bf f} = [f_1,\dots,f_i,\dots ]$ is the pooled feature for that cell.
Note that  those overlapped pixels in neighboring patches generally have different pixel
values after applying contrast normalization.
A comparison of these two pooling methods are evaluated in Section~\ref{SEC:model}.

Images (patches) of multiple scales have been widely used in computer vision
to incorporate information at different scales.
For our method, we have also found that  extracting multi-scale patches almost
always improves the classification accuracy.
Therefore, we have used image patches of different sizes (\eg, $4 \times 4, 6 \times 6$ or $8 \times 8$ pixels)
and concatenated all the pooled features on multi-scale patches to form the final feature vector.

Thus far, we have discussed how to extract the final features. In practice, we find that
standardizing the extracted features before training the linear classifier
almost always
leads to superior results (see Table \ref{Tab:normalization}).
To perform feature standardization, we subtract the mean and divide by the standard
deviation of each feature in the training set, following \cite{coates2011importance}. 

\subsection{Linear multi-class classification}

In this work, we feed the extracted features to a linear classifier to recognise the probe face image.
A simple ridge regression based multi-class classifier was used in \cite{gong2011comparing}.
We use the same linear classifier for its computational efficiency.
The ridge regression classifier has a closed-form solution, which makes the training
even faster  than the
specialized linear support vector machine (SVM) solver
{\sc Liblinear}  \cite{liblinear08}.
Despite its simplicity, the classification performance of this ridge regression approach
is on par with linear SVM \cite{gong2011comparing}.
Another benefit of this classifier is that, compared to the one-versus-all SVM, it only needs
to compute the classification matrix $\BW$ once.

We summarize our method in Algorithm~\ref{Alg:pooling}. It is very simple and the entire feature extraction
algorithm can be implemented within 20 lines of Matlab code (see the supplementary document).

\begin{algorithm}[t]
\small
\SetAlgoLined
\SetAlgoVlined
\KwIn{Training images and test samples}

Define the image patch sizes and pooling pyramids\;
\For{each patch size}{
$ \cdot $\, Densely extract local patches for both training and testing data\;
$ \cdot $\, Conduct contrast normalization\;
$ \cdot $\, Apply dimensionality reduction on the patches by using PCA\;
$ \cdot $\, Apply polarity splitting on the pooled features\;
$ \cdot $\, Perform spatial pooling over the pyramid grids\;
}

Concatenate the obtained features with all patch sizes\;
Standardize the obtained feature data\;
Train a multi-class linear classifier\;

\KwOut{The class label of a test face using the trained classifier}
\caption{ Face recognition with spatial pooling of local patches}
\label{Alg:pooling}
\end{algorithm}

%
%

\section{Model selection}
\label{SEC:model}
In this section, we test the impact of the components of the proposed algorithms.
The key parameters are the pooling pyramid levels, and the multiple different patch
sizes. The evaluation is conducted on the LFW-a dataset \cite{LFW_a}, and all the images
are down-sampled to $64 \times 64$ pixels. The dataset's description can be found in Section
\ref{SEC:exp}. We set the number of training and testing samples per subject to 5 and 2,
respectively. All the results reported in this section are based on 5 independent data
splits.
We empirically reduce the dimension of image patches to 10 by PCA, which preserves about
90\% of the eigenvalue energy.


\subsection{Multi-level spatial pooling}

First, we examine the importance of pooling for face recognition.
For this purpose, we test different pooling pyramids: from the 1-level pyramid \{1\}
to a
maximum 8-level pyramid $\{1, 2, 4, 6, 8, 10, 12, 15\}$.
With this 8-level pyramid, pooling is performed on regular grids of $ 1 \times 1$, $ 2 \times 2$,
     \dots, $ 15 \times 15$ and the obtained pooled features are concatenated altogether.

We also test the two popular pooling
strategy: max-pooling and average pooling.
As the default setup, local patches of size $4 \times 4$ pixels
are densely extracted with stride step of 1 pixel on the input raw images.
Thus we in total  extract $3721$ local patches for each image at this setting.

\begin{figure}[]
\centering
\includegraphics[width=0.4\textwidth]{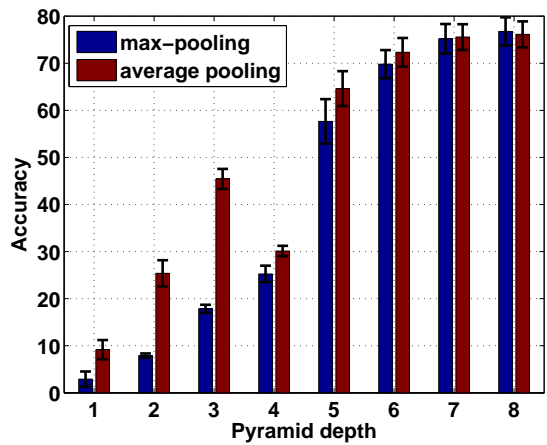}
\caption{Classification accuracy (\%) with varying the pooling pyramid levels.
     Here the 8-level pyramid is set to \{1, 2, 4, 6, 8, 10, 12, 15\}.
}
\label{Fig:pyramid}
\end{figure}

The recognition accuracies for different pyramid settings
are shown in Fig.~\ref{Fig:pyramid}. It is clear
that, for both max-pooling and average pooling,
the face recognition performance is significantly and almost consistently improved as the
number of pooling levels increases. Pooling effectively encodes multiple levels of spatial
information, which  appears extremely useful for classification. 
From this experiment, it seems that a 8-level pyramid is sufficient. {\em This is in contrast
to the SPM for generic image classification,
for which one usually does not observe significant performance improvement beyond
3 levels of pooling as shown in \cite{SPM2006}.}

Fig.~\ref{Fig:pyramid} also shows an interesting result that average pooling achieves much
better results than max-pooling when no more than 6 pooling levels are applied.
This might be due to the fact that with fewer pooling levels,
max-pooling has discarded too much information while for average pooling, all pixels in the
pooling cell contribute to the final result.

With the pyramid levels being
larger than 6, max pooling achieves close or slightly higher accuracy than average pooling.
This is very different to the results reported in generic image classification, where max pooling
used in the BOF models usually leads to superior performance than average pooling
with a pyramid of only 3 levels \cite{boureau2010learning,yang2009linear}.


\subsection{Dense extraction of local patches}

\begin{figure}[]
\centering
\includegraphics[width=0.4\textwidth]{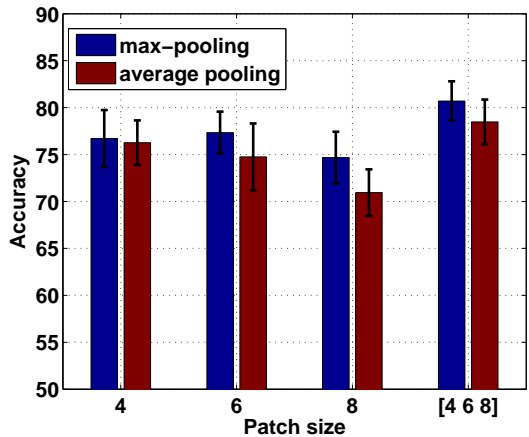}
\caption{Accuracy (\%) with varying the patch size. $ 4, 6, 8$ means
     patch sizes of $ 4 \times 4$, $ 6 \times 6$, $ 8 \times 8$ pixels.
     Here we use a 8-level pyramid for pooling. }
\label{Fig:rfsize}
\end{figure}

We then test the impact of patch sizes on the recognition accuracy.
We use different patch sizes of $\{4 \times 4, 6 \times 6, 8 \times 8 \}$ pixels,
   and combinations of them. The comparative
results are reported in Fig.~\ref{Fig:rfsize}. Here we can clearly see that smaller patch
sizes of $ 4 \times 4 $ and  $ 6 \times 6$ pixels
lead to better results than the patch size of $8 \times 8$. Similarly, smaller
receptive fields (patch sizes) are critical for achieving high performance in the
generic image classification with unsupervised feature learning
\cite{coates2011analysis}. It is also clear that the combination of the features extracted with
different patch sizes achieves significant performance improvements. Specifically, the
combination use of these three patch sizes obtains at least 3.4\% and 2.2\% accuracy gains
for max-pooling and average pooling, respectively.

All the previous evaluations are based on  dense extractions of local patches with a stride
step size of 1 pixel. Table~\ref{Tab:step} shows that if we use a larger
step size, the performance deteriorates.
Clearly, smaller steps are favoured.
Again, the same observation is made in  \cite{coates2011analysis} for generic image classification.

Through the evaluation of pooling pyramid depth and patch extraction, we can conclude that
\emph{multiple-level pooling and dense extraction of multi-scale image patches are
     critical for achieving high classification accuracies in
face recognition}. From the perspective of dimensionality, higher dimensional
features (due to the deeper pooling pyramid and multi-scale image patches)
leads to better performance.
This is consistent with the recent results reported in \cite{JianSun2013}.
The intuition  is that the high-dimensional feature vectors encode rich information for the
sequential linear classification. The high-dimensional vectors
may contain noises but  the linear classifier is able to select the relevant features
 and suppress the useless ones.

\setlength{\tabcolsep}{3pt}
\begin{table}
\centering
\caption{Impact of the step size on the
     recognition rate (\%) with different patch sizes.
     A 5-level pyramid of \{1, 2, 4, 6, 8\} is used here
     since the number of patches is only 16-by-16 at most when using a step size
     of 4 pixels.}
\begin{tabular}{cccc}
\hline
Step size & 1 pixel & 2 pixels & 4 pixels\\
\hline
$4 \times 4$ patch& $64.0 \pm 4.2$ &$50.3\pm 2.4$&$27.4\pm 1.9$\\
$6 \times 6$ patch& $66.5 \pm 3.6$ &$56.3\pm 3.1$&$35.3\pm 2.9$\\
$8 \times 8$ patch& $62.6 \pm 2.4$ &$60.6\pm 3.0$&$41.1 \pm 3.0$\\
\hline
\end{tabular}
\label{Tab:step}
\end{table}

\begin{table*}[t]
\centering
\caption{Accuracy (\%) of our method without the operation of contrast normalization,
     polarity splitting or feature standardization and with all of them.
           \No: this operation is not applied;
           \Yes: applied.
Note that the first two operations are performed  before pooling while standardization is conducted on the pooled features before classification.
The combination of patch sizes \{$4 \times 4$, $ 6 \times 6$, $ 8\times 8$\} is used.}
{
\begin{tabular}{r | cccc|cccc}
\hline
Operation & \multicolumn{4}{c|}{FERET} & \multicolumn{4}{c}{LFW-a}\\
\hline\hline
Contrast Norm. & \No& \Yes  & \Yes& \Yes&  \No& \Yes & \Yes& \Yes\\
Polarity splitting & \Yes& \No& \Yes&  \Yes& \Yes& \No& \Yes& \Yes\\
Standardization&\Yes & \Yes& \No & \Yes& \Yes & \Yes& \No & \Yes\\
\hline
Accuracy & $88.9 \pm 1.2$&$92.6 \pm 0.9$&$92.0 \pm 0.9$& $93.4 \pm 0.6$&$67.3 \pm 0.7$&$78.2 \pm 1.8$ &$73.0 \pm 2.8$ &$80.7 \pm 2.1$\\
\hline
\end{tabular}
}
\label{Tab:normalization}
\end{table*}

\subsection{Feature pre-processing}

In this section, we evaluate the effects of pre-processing steps before pooling features in our
Algorithm \ref{Alg:pooling}: contrast normalization and polarity splitting.
The evaluation is conducted on both of the LFW-a and FERET datasets with the image size of $64 \times 64$ pixels.
See the  datasets' description  in Section~\ref{SEC:exp}.
Table~\ref{Tab:normalization} reports the recognition accuracies of our method without each
of these operations and with all of them.

We can clearly see that, there is a significant benefit of contrast normalization on both datasets. Specifically,  improvements of 4.5\% and 13.4\% accuracies are achieved by the simple normalization step on the FERET and challenging LFW-a, respectively.
Polarity splitting of features also slightly improves the recognition performance, which increases the
accuracy by 0.8\% on FERET and 2.5\% on LFW-a.
We also find that standardizing the extracted features before training a linear
classifier almost always helps the final classification.
In this experiment, feature standardization improves the accuracy by 1.4\%  and 7.7\% on the two datasets, respectively.

\begin{figure}
\centering
\includegraphics[width=0.4\textwidth]{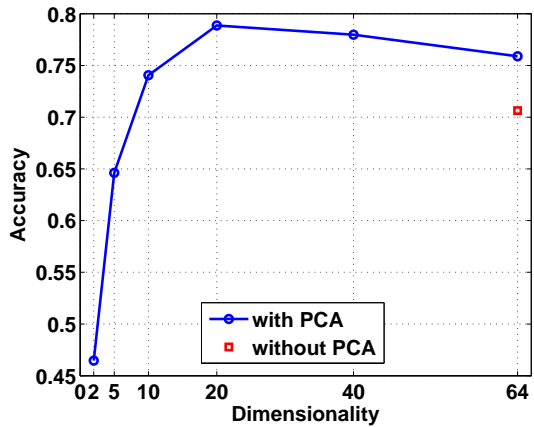}
\caption{PCA dimension versus accuracy using $8 \times 8$ patches. The result of  raw patches without applying PCA is marked as red. }
\label{Fig:pca-dim}
\end{figure}

Last, we will test the impact of the reduced dimensionality on the  recognition accuracy. We choose to  use the $8 \times 8$ patches. The results of PCA dimension versus accuracy are shown in Fig.~\ref{Fig:pca-dim}. The importance of applying PCA on raw patches is clearly shown. The original patches performs even worse than the 10-dim  PCA features.

All these feature processing steps are extremely simple.
However, as we can see,
     they play critical roles in practical face recognition tasks.
     The experimental results suggest that these operations should always be applied in our algorithm.


\section{Experimental results}
\label{SEC:exp}
In this section, we focus on the face identification problems. Our method are tested on four public facial image datasets, including AR \cite{AMM98}, CMU-PIE
\cite{PIE}, FERET \cite{FERET} and the challenging uncontrolled faces, LFW \cite{LFWTech}.
The compared face recognizers are the sparse representation based classifier (SRC) \cite{Wright09}),
robust sparse coding (RSC) \cite{RSC11}), superposed SRC (SSRC) \cite{dengdefense2013}),
local patch based Volterra \cite{kumar2012trainable} and multi-scale patch based MSPCRC
\cite{zhu2012multi}. 
These methods are reported to achieve the state-of-the-art results on the face identification problems in most recent years.
In section \ref{SEC:sspp}, we also compare our method to the popular generic image classification method locality-constrained linear coding (LLC \cite{LLC2010}) with multi-level spatial pyramid pooling.

As the authors recommend, we set the parameter $\lambda$ to 0.001,
0.001, 0.005 and 0.001 for SRC, RSC, SSRC and MSPCRC, respectively.
For Volterra \cite{kumar2012trainable},
we use the linear kernel since it always achieves better results than the quadratic one.
For RSC, Volterra and MSPCRC, we use the implementation code provided by the original authors.
In addition, we also compare with several recent published results.

According to the evaluation in the last section, for our two methods, we use max-pooling
with a 8-level pooling pyramid \{1, 2, 4, 6, 8, 10, 12, 15\}
and the combination of patch
sizes $\{4 \times 4, 6 \times 6, 8 \times 8\}$ pixels as the default setting\footnote{For images of size $32 \times 32$ (or smaller), we
only use one small patch size $\{4 \times 4\}$ due to the numerical problem with large-size
patches. }.
All of the patches are transformed to a 10D by applying PCA.

\subsection{AR}
The AR dataset \cite{AMM98} consists of over 4000 facial images from 126 subjects (70 men
and 56 women). For each subject 26 facial images were taken in two separate sessions.
The images exhibit a number of variations including various facial expressions (neutral,
smile, anger, and scream), illuminations (left light on, right light on and all side
lights on), and occlusion by sunglasses and scarves. Of the 126 subjects available 100
have been selected in our experiments (50 males and 50 females).

Since AR has been used very often  to evaluate face recognition algorithms,
we compare our method to several published state-of-the-art results on this dataset with four different situations.
For the `All' situation, all the 13
samples in the first session are used for training and the other 13 images in the second session for test.
For the `clean' situation, the 7 samples in each session with only illumination and expression changes are used. For the `sunglasses' or `scarf' situation, 8 clean samples from two sessions are for training and 2 images with sunglasses or scarf disguises are for test. All the images are
 resized to $64 \times 64$ pixels.
The  results are reported in Table~\ref{Tab:AR}.

It is clear that our method achieves better results than all other algorithms in all of the four situations. Specifically, our method outperforms the best of all other methods by 3.3\%, 1.0\%, 1.5\%, and 0.1\% in the `clean',`sunglasses', `scarf' and `All' situation, respectively. In particular, our algorithm obtains a perfect recognition with the sunglasses disguise.
{ To  our knowledge, ours is the  best reported result in this scenario.}

\begin{table}
\centering
\caption{Comparison with recent state-of-the-art results (\%) on AR with four different settings.
     For the first six methods, we quote the best reported results from the corresponding published papers, with the same experiment setting.}
\begin{tabular}{r | cccc}
\hline
Situation & Clean   & Sunglasses  & Scarves& All   \\
\hline\hline
SRC \cite{Wright09}&92.9&87.0&57.5&-\\
RSC \cite{RSC11} & 96.0  & 99.0 & 97.0& -\\
DRDL \cite{ma2012sparse} & 95.0 & - &- &-\\
$L_2$ \cite{Javen2011} & -  & 78.5 & 79.5& 95.9\\
SSRC \cite{dengdefense2013} & -  & 90.9 &90.9& 98.6\\
FDDL \cite{yang2011fisher} & 92.0 & - &- &-\\
Volterra &90.9&96.1&92.1&87.5\\
Ours & {\bf 99.3}  &{\bf 100.0}&{\bf 98.5}& {\bf 98.7}\\
\hline
\end{tabular}
\label{Tab:AR}
\end{table}

\begin{figure}
\centering
\includegraphics[width=0.4\textwidth]{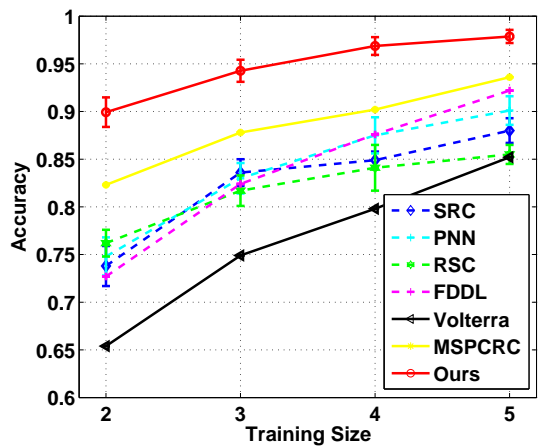}
\caption{Recognition accuracy on AR with images resized $32 \times 32$ pixels.
     Results are based on 20 independent runs.
     The results for PNN (patch based nearest neighbour)  \cite{kumar2011maximizing},
     Volterra and MSPCRC are quoted from \cite{zhu2012multi} with identical experiment settings.
     Again, ours outperform previous best results by a large margin.
}
\label{Fig:AR}
\end{figure}

Following the settings in \cite{zhu2012multi},
we further test the small-training-size problem on this dataset.
Among the samples with only illumination and expression changes in each class,
      we randomly select 2 to 5 images of the first session for training and 3 of the
second session for test. All the images are resized to $32 \times 32$ pixles. We independently
run the experiments for 20 times and report the average results in Fig.~\ref{Fig:AR}. We can
see that our method outperforms all other methods by large margins on this small-training-size problem.
The local patch based PNN and Voltera methods achieve inferior results to the multi-scale
based MSPCRC. The holistic representation based methods such as
SRC \cite{Wright09}, RSC \cite{RSC11} and
FDDL \cite{yang2011fisher} do not perform as well as the local patches
based MSPCRC, either. The superior performance of our method and MSPCRC conforms the importance
of extracting features based on multi-scale local patches.

\subsection{CMU-PIE}
The CMU-PIE dataset \cite{PIE} is widely used for face recognition. We use this dataset to compare several popular feature extraction methods (including Eigenfaces \cite{Turk91}, Fisherfaces \cite{FisherLDA}, Laplacianfaces \cite{Laplacianfaces05} and its orthogonal variation OLAP \cite{cai2006orthogonal}) and more recent
methods such as
SRC, MSPCRC, Volterra and the kernel ridge regression KRR \cite{KRR07}.

%
%
%

In our experiments, we use the data and settings provided by the authors of \cite{cai2006orthogonal}. The data includes 170 samples  each subject from five near frontal poses (C05, C07, C09, C27, C29).  From 2 to 30 samples are used for training and all the rest are for testing. Table \ref{Tab:PIE} compares the mean results of our method and several published results based on 10 independent runs.

Table \ref{Tab:PIE} clearly demonstrates the advantage of the proposed method.
In particular, {\em our method achieves 72.2\%  and 99.7\% accuracy with  2 and 30 training samples, which are higher than those of all other methods by about 15.5\% and 1.4\%, respectively}.
     Volterra performs the second best on this dataset. The recent state-of-the-art algorithms
SRC, RSC and MSPCRC do not show better performance than
OLAP on this relatively simple dataset.

\begin{table}
\centering
\caption{Comparison of average accuracy (\%) on CMU-PIE
     with varying training size.
          For the first six methods,
          we quote the best reported results from the cited papers.
          SRC, MSPCRC and our methods are conducted on images of
          $64 \times 64$ pixels.
}
\begin{tabular}{r | ccccc}
\hline
Training size & 2  & 5 & 10 & 20 & 30 \\
\hline\hline
Eigenfaces \cite{cai2006orthogonal} &- & 30.1&44.3 &61.9 &72.1 \\
Fisherfaces \cite{cai2006orthogonal}&-&68.5&77.6&84.6&92.2\\
Laplacianfaces \cite{cai2006orthogonal}& - &69.2 &78.9 &85.9 & 92.8\\
OLAP \cite{cai2006orthogonal} & -&78.6 & 88.6&93.5&95.2\\
KRR \cite{KRR07} &- & 73.6 &86.9&94.0&96.0\\
Volterra \cite{kumar2012trainable}& 57.0 & 80.8& 90.9&95.4 &98.3\\
SRC&46.0 &72.3 &85.7   &93.1 &95.1 \\
RSC&43.7 & 72.3 & 85.4& 92.5&94.5\\
MSPCRC&51.5  &  74.1  &  87.4   &  93.2  &  94.9     \\
Ours & {\bf 72.2} & {\bf 94.4}&{\bf 98.5 }&{\bf 99.5}&{\bf 99.7}\\
\hline
\end{tabular}
\label{Tab:PIE}
\end{table}



\subsection{FERET}
The FERET dataset \cite{FERET} is an widely-used standard face recognition benchmark set
provided by DARPA. Since the images of FERET dataset are acquired in multiple sessions
during several years, they have complex intraclass variability.
We use a subset of FERET which includes 200 subjects.
Each individual contains 7 samples, exhibiting
 facial expressions, illumination changes and up to 25 degrees of pose variations. It is
composed of the images whose names are marked with `ba', `bj', `bk', `be', `bf', `bd' and
`bg'. The images are cropped and resized to $80 \times 80$ pixels \cite{yang2003combined}. We
randomly choose 5 samples from each class for training and the remaining 2 samples for testing.
The mean results of 10 independent runs with images down-sampled to $32 \times 32$  and $64 \times 64$
pixels  are reported in Table~\ref{Tab:FERET}.

We can clearly see that our method achieves the highest recognition rate. In particular, with image size of $64 \times 64$ pixels,
{\em  our method obtains an accuracy of 93.4\%,
     which is much higher than the second best (82.1\% of SSRC \cite{dengdefense2013}) by 11.3\%.}
     RSC achieves similar results with SRC. The patch based MSPCRC and Volterra do not perform well,
   which may be because they are incapable to cope with the pose variations in this dataset.

\begin{table}
\caption{Average accuracy (\%) on FERET using 5 training samples each class with images of $32 \times 32$ and $64 \times 64$
     pixels. Results are based on 10 independent runs. }
\centering
\begin{tabular}{ccccccc}
\hline
Method  &MSPCRC & Volterra& SRC &  RSC&SSRC   & Ours\\
\hline
$32 \times 32$ pixels &$41.2 \pm 1.5$ &$46.2 \pm 1.6$ & $75.3 \pm 1.7$& $73.3 \pm 1.8$& $84.1 \pm 1.5$& $ \bf 87.0 \pm 1.2$\\
$64 \times 64$ pixels &$42.3 \pm 2.8$ &$50.4 \pm 2.3$& $73.8 \pm 2.0$&$73.4 \pm 2.6$&$82.1 \pm 1.6$&$\bf 93.4 \pm 0.6$\\
\hline
\end{tabular}
\label{Tab:FERET}
\end{table}


\subsection{LFW-a}
\begin{figure}
\centering
\includegraphics[width=0.5\textwidth]{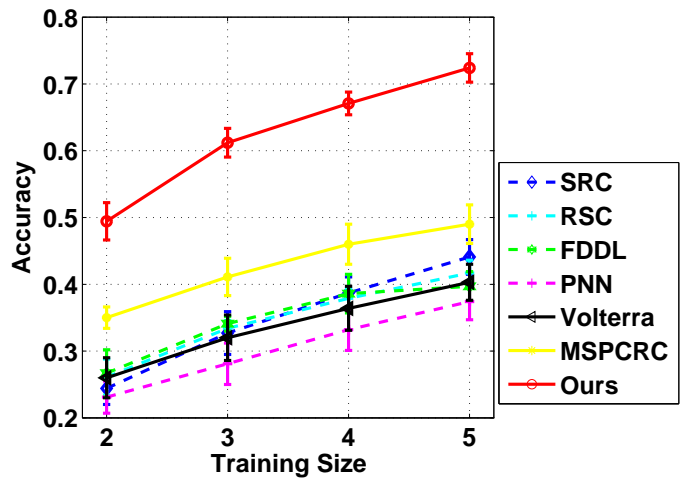}
\caption{Recognition accuracy on LFW with downsampled $32 \times 32$ images.
     Results are based on 20 independent runs. The results for PNN, Volterra and MSPCRC are quoted from \cite{zhu2012multi} with identical settings.}
\label{Fig:LFW}
\end{figure}

%
%
Following the settings in \cite{zhu2012multi},
     here we use LFW-a, an aligned version of LFW using commercial face alignment software \cite{LFW_a}.
     A subset of LFW-a including 158 subjects with each subject more than 10 samples are used. The images are cropped to $121 \times 121$ pixels.
      From    2 to 5 samples are randomly selected for training and another 2 samples for
testing. For a fair comparison, all the images are finally resized to $32 \times 32$ pixels
as  in \cite{zhu2012multi}. Fig.~\ref{Fig:LFW} lists  the average results of 20 independent runs.

Consistent with the previous results, {\em our method outperforms all other algorithms by even
larger gaps on this very challenging dataset.} 
 The representation based methods SRC, RSC and FDDL do not show significant differences and perform sightly better than the patch based nearest neighbour classifier PNN.
 We obtain higher accuracies than the second best
method MSPCRC by  $14.4\% \sim 23.4\%$. Actually if we use a larger image size $64 \times 64$, the accuracy of our method will increase to 80.7\% , while MSPCRC is not able to improve the accuracy (47.7\%) in this case. In our experiments, we find that for our method,
a large image size almost always achieves better results than using a smaller one.


\subsection{Comparison with local feature or patch based methods}
\label{SEC:sspp}

Local feature (or patch) based face recognition methods tends to show superior  results to the holistic feature (or patch) based ones, especially on the single sample per person (SSPP) problem. Since only one
training sample for each person is registered in the training data, many holistic methods (\eg, SRC) tend to fail in this case.
Due to this, we further evaluate our method against several representing local feature based methods, which were reported to achieve good results on SSPP problems. 
These methods include Histogram of Gabor Phase Patterns (HGPP \cite{zhang2007histogram}), 
Zou's method \cite{zou2007comparative} and  Tan's method \cite{tan2007fusing}. We also  compare our method with the popular BoF  approach locality-constrained linear coding (LLC \cite{LLC2010}), which can be regarded as an efficient version of the ScSPM method \cite{yang2009linear}. The dictionary is generated by K-means on $6 \times 6$ patches with 1024 basis. We set the regularization parameter $\lambda = 0.001$ and the pyramid as [1, 2, 4].
For fair comparison, contrast normalization and polarity splitting are performed for LLC as in our method, which are shown to improve the results.

We conduct the comparison on the FERET dataset with standard protocols.  Taking the subset Fa with one image for each of 1196 subjects as a gallery, Fb (expression variations) and Fc (illumination variations), and DupI and DupII (age variations) are taken as the probe subsets.
Each image is cropped based on the locations of eyes and then normalized to $150 \times 130$ pixels. From Table~\ref{Tab:sspp}, we can see that our approach achieves very competitive results with the local feature based methods, even if ours is much simpler. 

Compared to the BoF model LLC, our method performs much better by excluding the dictionary training and feature encoding procedures. This is mainly due to the fact that our method can involve much more levels of pooling than LLC. The encoding procedure (with  hundreds or thousands of dictionary basis) makes the dimension of the learned feature too large to conduct high-level pooling. Spatial pooling are shown to be more critical than  feature encoding, at least on the face recognition problems. 
From Table~\ref{Tab:LFW_Dim}, similar results are observed on face recognition problems with more training samples. The proposed pooling method outperforms the BoF model LLC by 14.7\% and 7.2\% with  $32 \times 32$ and $64 \times 64$ images, respectively.

\begin{table}
\centering
\caption{Accuracy (\%) on FERET with standard protocols.}
\begin{tabular}{c|cccc}
\hline
Method & Fb (expression) & Fc (illumination)& DupI (aging) & DupII (aging)\\
\hline
\hline
HGPP \cite{zhang2007histogram} & 97.5 & \textbf{99.5} & 79.5 & 77.8\\
Zou's method \cite{zou2007comparative} & 99.5 & \textbf{99.5} & 85.0 & 79.5\\
Tan's method \cite{tan2007fusing}& 98.0 & 98.0 & \textbf{90.0} & 85.0\\
LLC & 99.4 & 96.9 & 82.3 & 77.8\\
Ours &\textbf{99.7} & 98.5&\textbf{89.1} & \textbf{85.5}\\
\hline
\end{tabular}
\label{Tab:sspp}
\end{table}

\begin{table}
\centering
\caption{Recognition accuracy (\%) of local patch based methods on LFW-a with 5 training samples each class.}
\begin{tabular}{r  cc}
\hline
Method & $32 \times 32$ pixels& $64 \times 64$ pixels\\
\hline
LLC & $57.7 \pm 1.8$& $72.5 \pm 2.1$\\
Ours & ${\bf72.4 \pm 2.1}$ & ${\bf 80.7 \pm 2.1}$  \\
\hline
\end{tabular}
\label{Tab:LFW_Dim}
\end{table}

\section{Conclusion}
In this work, we have proposed an extremely simple yet effective face recognition method
based on
spatial pooling of local patches, which has two major components: local patch extraction
and spatial pooling.
We show that \textit{dense extraction} and \textit{multi-level pooling} are
critical for local patch based face recognition. The extracted facial features of different
subjects by our method demonstrate strong characteristics of clusters without using the
label information.


Despite its simplicity, the proposed method surprisingly outperforms many recent state-of-the-art methods (both holistic and local patch based)
by large margins on four public benchmark datasets (\eg, over 20\% on LFW-a).
The proposed method also works  better than the BOF model LLC (with the additional feature encoding step) on face recognition problems.

An extension is to apply the simple method on generic image classification problems.
Compared to the popular image  feature learning and encoding methods, the proposed method has no need for 
dictionary training or feature encoding. We plan to explore this possibility in the future work.


\vspace{1cm}
\section*{Appendix: MATLAB code of our feature extraction method}
\vspace{0.5cm}

\verbinput{fea_pooling.m}

\end{document}